\documentclass[runningheads]{llncs}

\usepackage{eccv}

\usepackage{eccvabbrv}

\usepackage{graphicx}
\usepackage{booktabs}

\usepackage[accsupp]{axessibility}  

\usepackage{hyperref}
\usepackage{wrapfig}
\usepackage{booktabs}
\usepackage{subcaption}
\usepackage{pgfplots}
\pgfplotsset{compat=1.18}
\usepackage{orcidlink}

\begin{document}

\title{vGamba: Attentive State Space Bottleneck for efficient Long-range Dependencies in Visual Recognition} 

\titlerunning{vGamba: visual bottleneck}

\author{Yunusa Haruna\inst{1}\orcidlink{} \and
Adamu Lawan\inst{1,2,3}\orcidlink{} \and
Shamsuddeen Hassan Muhammad\inst{4}\orcidlink{} \and
Jiaquan Zhang\inst{2,3}\orcidlink{5} \and
Chaoning Zhang\inst{3}\orcidlink{5}
}

\authorrunning{Y.~Haruna et al.}

\institute{NewraLab, Suzhou, China \and
Beihang University, Beijing, China \and
Beijing GoerTek Alpha Labs, Beijing, China \and
Imperial College London, UK \and
University of Electronic Science and Technology of China\\}

\maketitle

\begin{abstract}
Capturing long-range dependencies (LRD) efficiently is a core challenge in visual recognition, and state-space models (SSMs) have recently emerged as a promising alternative to self-attention for addressing it. However, adapting SSMs into CNN-based bottlenecks remains challenging, as existing approaches require complex pre-processing and multiple SSM replicas per block, limiting their practicality. We propose vGamba, a hybrid vision backbone that replaces the standard bottleneck convolution with a single lightweight SSM block, the Gamba cell, which incorporates 2D positional awareness and an attentive spatial context (ASC) module for efficient LRD modeling. Results on diverse downstream vision tasks demonstrate competitive accuracy against SSM-based models such as VMamba and ViM, while achieving significantly improved computation and memory efficiency over Bottleneck Transformer (BotNet). For example, at $2048 \times 2048$ resolution, vGamba is $2.07\times$ faster than BotNet and reduces peak GPU memory by $93.8\%$ ($1.03~\text{GB}$ vs.\ $16.78~\text{GB}$), scaling near-linearly with resolution comparable to ResNet-50. These results demonstrate that Gamba Bottleneck effectively overcomes the memory and compute constraints of BotNet global modeling, establishing it as a practical and scalable backbone for high-resolution vision tasks.
\keywords{Attention Mechanism \and Bottleneck \and CNN \and SSM}
\end{abstract}
\section{Introduction}
\label{sec:intro}

\begin{figure}[t]
\centering

\begin{minipage}{0.48\textwidth}
    \centering
    \includegraphics[width=\linewidth,trim=0 0 0 0,clip]{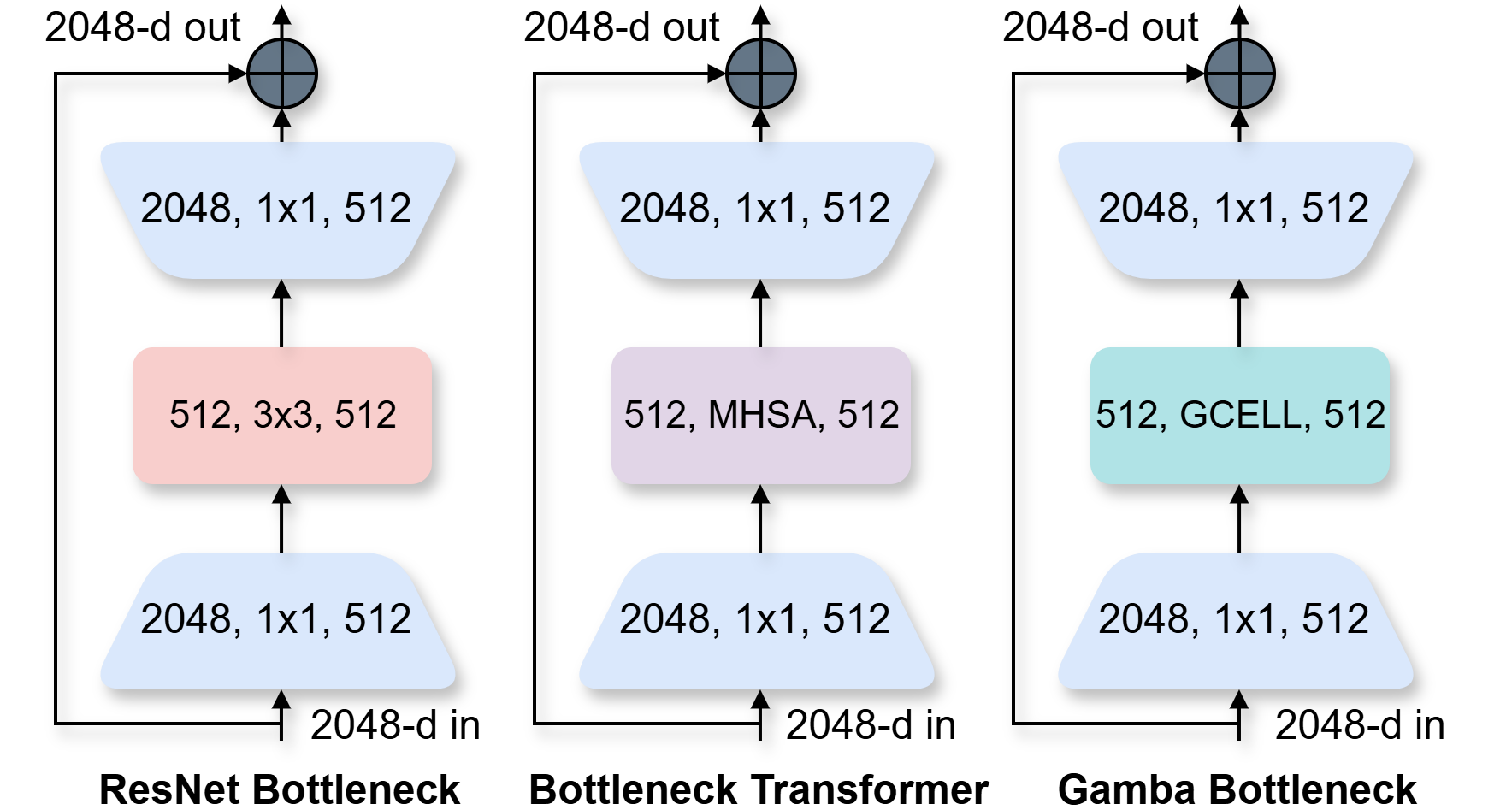}
\caption{Bottleneck evolution: CNN, Transformer and SSM. BoTNet replaces the $3\times3$ convolution with MHSA while vGamba replaces it with GCell. At $2048\times2048$ resolution, vGamba is $2.07\times$ faster than BoTNet and reduces peak GPU memory by $93.8\%$ (Sec. ~\ref{sec:scaling}).}
    \label{fig:cnn_vit}
\end{minipage}
\hfill
\begin{minipage}{0.48\textwidth}
    \centering
    \begin{tikzpicture}
    \begin{axis}[
        ybar,
        bar width=9pt,
        width=\linewidth,
        height=0.6\linewidth,
        ymin=0,
        ylabel={Value},
        ylabel style={font=\scriptsize},           
        symbolic x coords={ResNet, BotNet, vGamba},
        xtick=data,
        tick label style={font=\scriptsize},       
        nodes near coords,
        nodes near coords style={font=\tiny},
        grid=major,
        legend style={font=\tiny, at={(0.5,-0.28)}, anchor=north, legend columns=-1},
        enlarge x limits=0.3,
        major grid style={gray!30}
    ]
    \addplot[fill=blue!70!white] coordinates {(ResNet,25.6) (BotNet,20.8) (vGamba,19.4)};
    \addplot[fill=red!70!white] coordinates {(ResNet,3.9) (BotNet,4.1) (vGamba,3.8)};
    \addplot[fill=gray!70!white] coordinates {(ResNet,76.2) (BotNet,78.3) (vGamba,81.1)};
    \legend{Param (M), FLOPs (G), Top-1k (\%)}
    \end{axis}
    \end{tikzpicture}
    \captionof{figure}{Quantitative comparison of bottleneck designs. vGamba gains $+2.8\%$ Top-1 over BoTNet with $1.4$M fewer parameters and $0.3$G lower FLOPs, showing that the proposed GCell achieves better representational capacity at lower computational cost (Sec. \ref{sec:classification}).  }
    \label{fig:bottleneck_bar}
\end{minipage}
\vspace{-20pt}
\end{figure}

Modeling long-range dependencies (LRD) is essential in visual recognition for interpreting complex scenes where objects span large spatial regions, such as in instance segmentation or aerial imagery. Traditional convolutional neural network (CNN) backbones have dominated the field \cite{szegedy2015going,he2016deep,chollet2017xception,tan2019efficientnet,radosavovic2020designing}, and while their hierarchical structure allows effective feature extraction, their strong inductive bias and local receptive fields (RF) limit their ability to model LRD. To mitigate this, larger kernels and dilated convolutions have been used to expand the RF; however, larger kernels are computationally expensive, and dilation causes spatial discontinuities that reduce smooth LRD modeling \cite{ding2022scaling,yu2017dilated}. The evolution of bottleneck designs from CNNs to Transformers and more recently to State-Space Models (SSMs) is illustrated in Fig.~\ref{fig:cnn_vit}. 
While BoTNet improves LRD, its computational overhead is higher. A quantitative comparison of representative bottleneck designs in terms of parameters, FLOPs and accuracy is provided in Fig.~\ref{fig:bottleneck_bar}, highlighting the trade-offs between efficiency and performance.

\begin{figure*}[htbp]
  \centering
  \renewcommand{\arraystretch}{1.0}
  \setlength{\tabcolsep}{2pt}
  \begin{tabular}{cccc}
    \includegraphics[width=0.70\textwidth,height=0.17\textwidth]{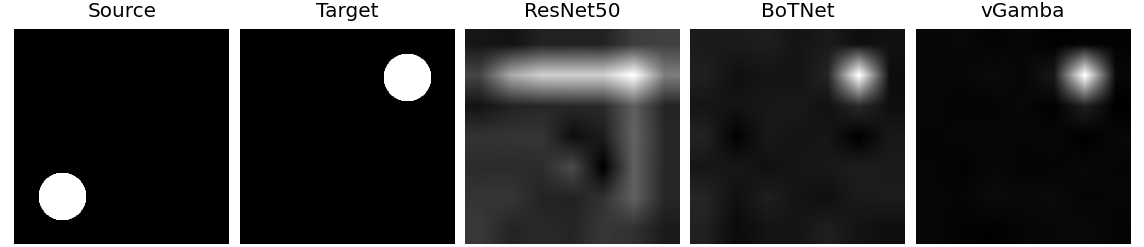} &
    \includegraphics[width=0.25\textwidth,height=0.15\textwidth]{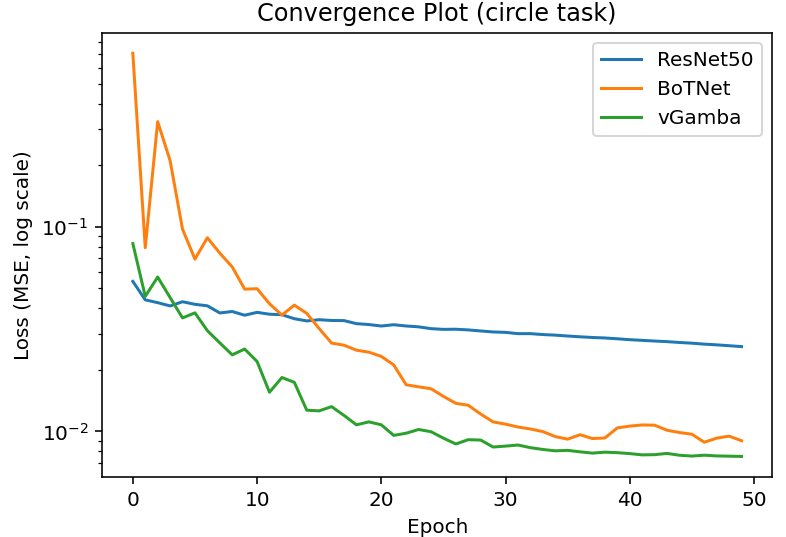} \\
    \includegraphics[width=0.70\textwidth,height=0.17\textwidth]{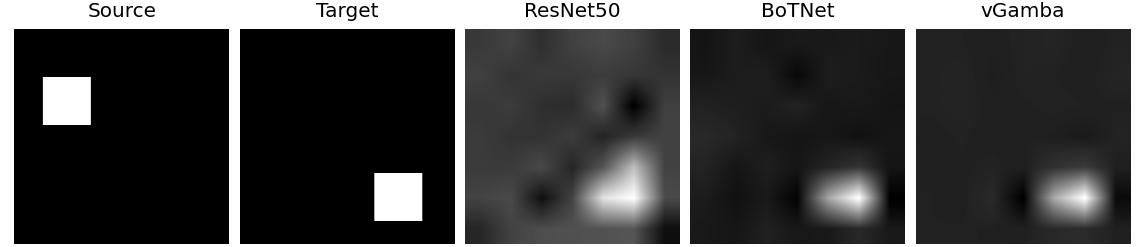} &
    \includegraphics[width=0.25\textwidth,height=0.15\textwidth]{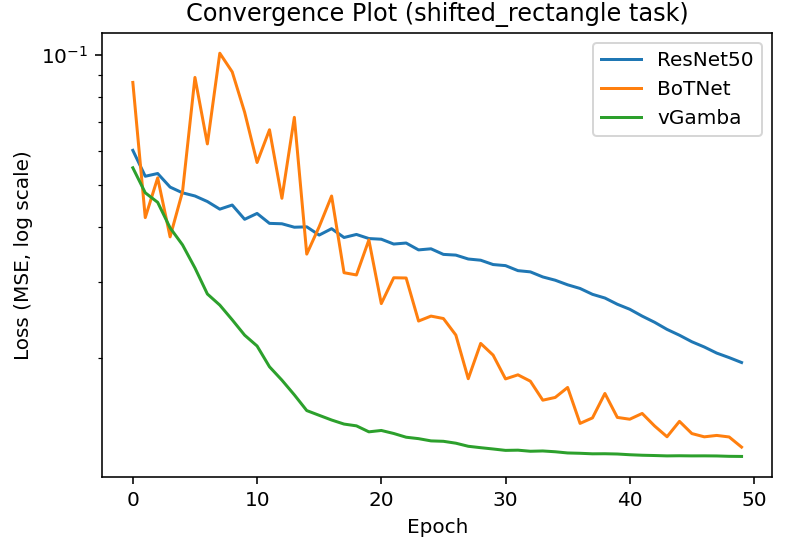} \\

  \end{tabular}

\caption{To test the models' ability to capture LRD, we transport information between distant spatial diagonal locations, placing the source and target in opposite corners using different shapes. We compare three bottleneck architectures ResNet-50 (CNN), BotNet (Transformer), and vGamba (SSM) all trained under identical settings (AdamW, lr 0.001, 50 epochs, $224^2$, MSE). ResNet-50 shows an almost straight-line convergence curve, indicating limited capacity to model LRD. BotNet achieves better LRD capture than ResNet-50, yet at a higher computational cost due to its quadratic SA complexity. In contrast, vGamba converges faster and more smoothly, producing more accurate reconstructions than both ResNet-50 and BotNet, effectively capturing LRD and spatial transport patterns at lower computational cost.}
\label{fig:long_range}

\end{figure*}

Consequently, Vision Transformers (ViT) \cite{dosovitskiy2021image}, \cite{haruna2025exploring} have gained popularity for their ability to capture LRD using self-attention (SA) mechanisms, which allow them to model global context more effectively than CNN, as shown in the spatial transport and convergence results in  Fig.~\ref{fig:long_range}. However, the computational cost of SA scales quadratically with sequence length, making ViT computationally expensive, especially for high-resolution images and real-time applications. Various techniques, such as low-rank factorization \cite{yang2024efficient}, linear approximation \cite{ma2021luna, song2021ufo}, and sparse attention \cite{chen2023sparsevit}, have been proposed to reduce computational costs while preserving performance. However, these approaches often involve trade-offs between efficiency and effectiveness.

State-space models (SSM)~\cite{gu2022s4} have recently emerged as an alternative to SA mechanisms, offering a more efficient way to model LRD. By capturing sequential dynamics in a compact form, SSMs reduce computational overhead while maintaining competitive performance. Mamba \cite{gu2023mamba}, a selective SSM, was introduced to further optimize efficiency, achieving up to $5\times$ higher throughput on sequence modeling benchmarks by selectively retaining relevant information and reducing redundant computations. However, adapting SSMs for vision tasks remains challenging, as it requires specific spatial pre-processing. For instance, VMamba \cite{liu2025vmamba} employs cross-scan and merging strategies requiring four SSM replicas per block, while ViM \cite{zhu2024vision} introduces bidirectional SSMs requiring two replicas per forward pass. Although effective, these designs are not suited as drop-in replacements for the single $3\times3$ convolution in a CNN bottleneck, as the added replicas increase architectural complexity and memory overhead. This motivates the need for a lightweight, single-block SSM design that integrates directly into standard CNN pipelines. See Fig. ~\ref{fig:comparison_architecture} for architectural illustrations.

To this end, we introduce vGamba, inspired by the Transformer bottleneck\cite{srinivas2021bottleneck}, we design a lightweight bottleneck module that integrates into the final layer of ResNet to efficiently capture LRD through a single SSM block. vGamba is a hybrid vision backbone built upon the Gamba Block, a compact structure derived from CNN bottlenecks. The Gamba cell extends Mamba for vision tasks by adapting it to 2D spatial structures and integrating ASC for enhanced spatial awareness. By combining these components, vGamba improves representational efficiency and contextual understanding, serving as an efficient alternative to conventional vision backbones. The contributions of this work are stated below:

\begin{itemize}
    \item We identify that existing SSM-based vision models require complex 
    pre-processing and multi-SSM stacking, making them unsuitable as 
    drop-in replacements for the $3\times3$ convolution in CNN bottlenecks. 
    We address this with the Gamba bottleneck, a lightweight single-block 
    SSM design with 2D positional awareness and an ASC module for efficient 
    LRD modeling.

    \item vGamba achieves competitive accuracy against SSM-based models 
    including VMamba and ViM across classification, detection, and 
    segmentation, while using fewer parameters and lower FLOPs 
    (Sec.~\ref{sec:experiments}).

    \item vGamba demonstrates near-CNN-level scaling efficiency: at 
    $2048\times2048$ resolution it is $2.07\times$ faster than BoTNet 
    and reduces peak GPU memory by $93.8\%$, making it practical for 
    high-resolution vision tasks (Sec.~\ref{sec:scaling}).
\end{itemize}

\section{Related Work}
\label{sec:relatedwork}
\begin{figure*}[htbp]
    \centering
    \includegraphics[width=\textwidth]{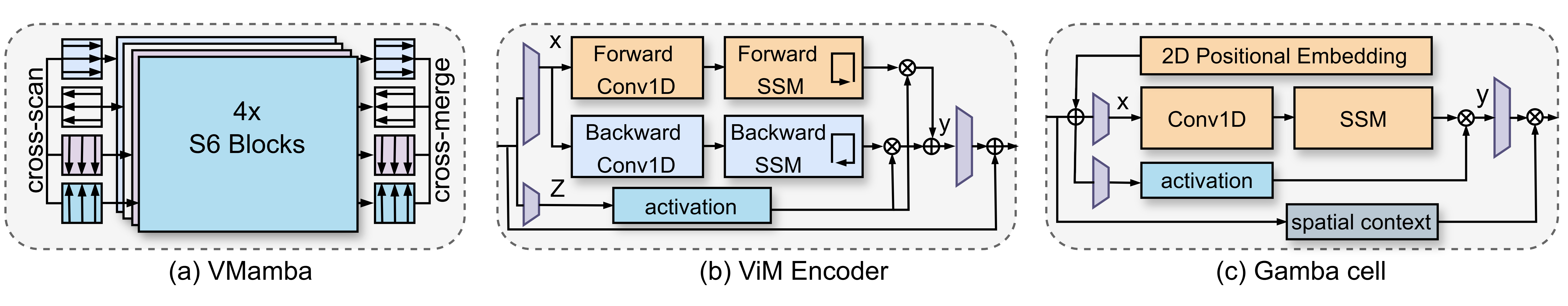}

\caption{Schematic comparison of the proposed Gamba cell (c) against (a) VMamba \cite{liu2025vmamba} and (b) ViM \cite{zhu2024vision} as drop-in replacements for the $3\times3$ convolution in the final ResNet bottleneck stage. VMamba requires four SSM replicas per block and ViM requires two, incurring higher complexity and FLOPs. Under identical settings, vGamba-B outperforms VMamba-B by $+1.8\%$ and ViM-B by $+0.6\%$ in Top-1 accuracy, while using $6.11$M and $2.60$M fewer parameters respectively, demonstrating the best efficiency-accuracy trade-off with a single SSM block (Appendix A.3.).}

    \label{fig:comparison_architecture}
\end{figure*}
Efforts to capture LRD in visual data have evolved across several paradigms. Early CNN-based approaches focused on enlarging the RF by using larger kernels or dilated convolutions. For instance, the Global Convolutional Network~\cite{peng2017large} used large kernels to model LRD, while the Dilated Residual Network~\cite{yu2017dilated} leveraged dilation to expand the RF without increasing model depth or complexity. While both approaches are effective, the former increases computational cost, whereas the latter can introduce gridding artifacts. An alternative is the Non-Local Neural Network~\cite{wang2018non}, which mitigates this by computing responses as weighted sums over all positions, effectively capturing spatial–temporal LRD, but at high computational cost.

Building on this direction of global modeling, ViT~\cite{dosovitskiy2021image} modeled global context from the first layer, treating image patches as tokens and linking them through SA, effectively capturing LRD. However, their quadratic complexity and large data demands make them resource-intensive. Several works addressed this: EfficientFormer~\cite{yang2024efficient} reduced backpropagation cost via low-rank projections, while LUNA~\cite{ma2021luna} and UFO-ViT~\cite{song2021ufo} proposed linear variants. SparseViT~\cite{chen2023sparsevit} further improved efficiency by pruning inactive tokens. Despite these advances, SA-based models remain costly for high-resolution visual tasks.

More recently, Mamba~\cite{gu2023mamba} applied SSMs to sequence modeling, offering efficiency and scalability beyond traditional attention. Originally designed for auto-regressive tasks, its 1D structure posed challenges for 2D vision. VMamba~\cite{liu2025vmamba} extended Mamba with cross-scan and cross-merge operations to capture pixels in all directions, creating four replicas of the S6 blocks. ViM~\cite{zhu2024vision} introduced bidirectional SSMs, creating forward and backward blocks to better capture LRD in images, enabling effective visual representation learning. These adaptations demonstrate that SSMs can provide an efficient alternative to ViTs, but at the cost of architectural complexity. However, integrating SSMs into standard CNN pipelines remains challenging. Existing SSM-based models require heavy preprocessing such as cross-scan, cross-merge, or bidirectional replication, resulting in multiple SSM instances per block and increased architectural complexity, Fig.~\ref{fig:comparison_architecture}. 

To achieve a balance between efficiency and simplicity, it becomes essential to explore lightweight mechanisms that can embed sequence modeling within CNN backbones without architectural overhead. Inspired by Bottleneck Transformers~\cite{srinivas2021bottleneck}, which successfully incorporate MHSA into CNN, we aim to bring SSM into CNN pipelines in a similarly efficient and modular manner.

\section{Preliminaries}
\subsubsection{SSM} represents an input sequence $x(t) \in \mathbb{R}$ through a latent state $h(t) \in \mathbb{R}^N$ using the following continuous linear system:
\begin{equation}
\begin{aligned}
h'(t) &= A h(t) + B x(t), \\
y(t) &= C h(t) + D x(t),
\end{aligned}
\end{equation}
where $A \in \mathbb{R}^{N \times N}$ is the state transition matrix, $B \in \mathbb{R}^{N \times 1}$ maps the input to the state, $C \in \mathbb{R}^{1 \times N}$ maps the state to the output, and $D \in \mathbb{R}$ is a feedthrough term.

To apply SSMs in deep learning, the continuous system is discretized using the Zero-Order Hold (ZOH) method, which assumes the input remains constant over each time step $\Delta t$. The discretized system is defined as:
\begin{equation}
\hat{A} = e^{A \Delta t}, \quad 
\hat{B} = \left( \int_{0}^{\Delta t} e^{A \tau} d\tau \right) B,
\end{equation}
leading to the discrete recurrence:
\begin{equation}
\begin{aligned}
h_k &= \hat{A} h_{k-1} + \hat{B} x_k, \\
y_k &= C h_k + D x_k.
\end{aligned}
\label{eq:eqn3}
\end{equation}

Earlier deep SSM variants such as S4~\cite{gu2022s4} introduced efficient parameterizations and fast discretizations for long-range modeling. Mamba~\cite{gu2023mamba} extends these formulations by making the parameters $B$, $C$, and the discretizations scale input-dependent. This dynamic adaptation allows the model to selectively retain relevant information and efficiently model LRD, achieving strong scalability and efficiency beyond traditional attention mechanisms. However, Mamba's causal formulation is designed for sequential data, making direct application to 2D visual inputs non-trivial, a limitation addressed by the proposed Gamba cell (Sec.~\ref{sec:gambacell}).
\section{Method}
\begin{figure*}[htbp]
    \centering
    \includegraphics[width=\textwidth]{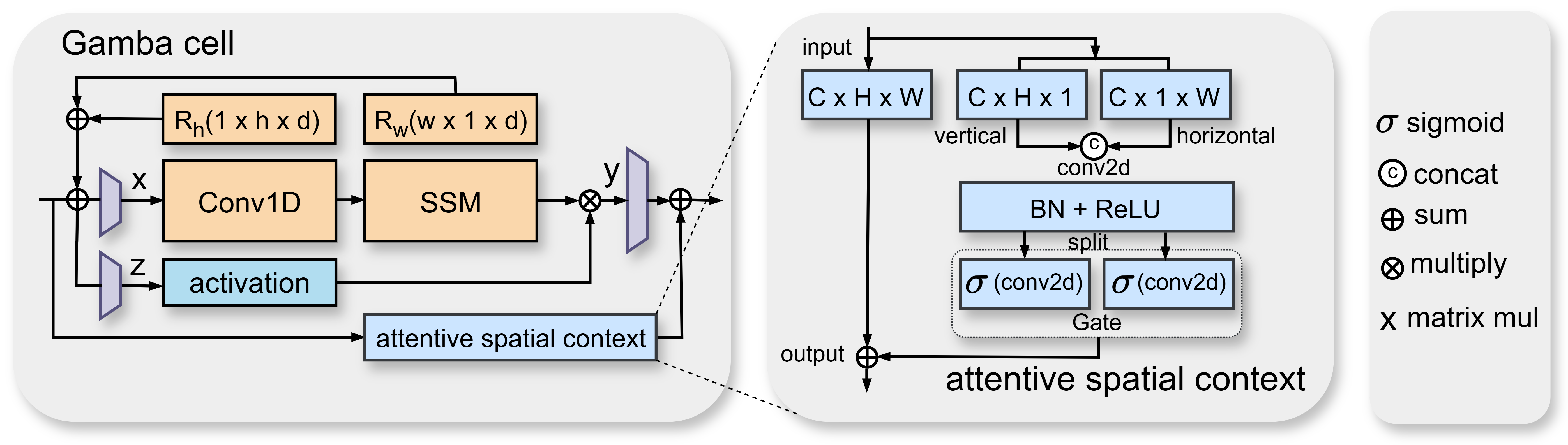}
    \caption{Gamba Bottleneck showing Gamba Cell and the ASC module.}
    \label{fig:architecture}
\end{figure*}

\subsubsection{Problem Statement.} Existing SSM adaptations for visual tasks, such as VMamba~\cite{liu2025vmamba} and ViM~\cite{zhu2024vision}, rely on cross-scan, cross-merge, or bidirectional operations to adapt 1D SSMs to 2D spatial structures, enabling LRD modeling. While effective at modeling global context, these designs require multiple SSM replicas per block, significantly increasing memory usage and architectural complexity, see Fig.~\ref{fig:comparison_architecture}. This makes them unsuitable as drop-in replacements for the standard $3\times3$ convolution in CNN bottlenecks, which are designed for efficiency, modularity, and simplicity.

\subsubsection{Goal.} Our objective is to adapt Mamba-style SSMs to fit directly into a CNN bottleneck, replacing the $3 \times 3$ convolution with a single forward SSM pass per block. This design aims to efficiently capture LRD within a standard CNN pipeline while maintaining low computational and memory overhead. By integrating SSMs in a modular and lightweight manner, we aim to surpass the effectiveness of traditional CNN and reduce the memory and computational demands of Bottleneck Transformers~\cite{srinivas2021bottleneck}, whose MHSA has quadratic complexity.

\subsection{Gamba Cell}
\label{sec:gambacell}
The original Mamba model relies on causal state transitions, where the $i$-th token depends only on tokens up to position $(i-1)$ (Eq.~\ref{eq:eqn3}). While this is well-suited for auto-regressive tasks, it introduces long-range decay in visual processing, weakening spatial dependencies over distance. Unlike VMamba~\cite{liu2025vmamba} and ViM~\cite{zhu2024vision}, which mitigate this by replicating the SSM block in multiple directions, the Gamba cell preserves the single causal SSM pass and compensates through 2D Relative Positional Embeddings (RPE) ~\cite{wu2021rethinking} and the ASC module, maintaining efficiency without sacrificing spatial context. The Gamba cell is shown in Fig.~\ref{fig:architecture}.

\noindent\textbf{Spatial Sequence Encoding.} Given an input feature $X \in \mathbb{R}^{B \times C \times H \times W}$, the Gamba cell proceeds as follows: (i) reshape $X$ to a sequence $X_{\text{seq}} \in \mathbb{R}^{B \times (HW) \times C}$ and inject 2D RPE, (ii) pass through the Mamba block $M$, (iii) apply the ASC module for spatial context refinement (Fig.~\ref{fig:ASC}), and (iv) reshape back to $X_{\text{out}} \in \mathbb{R}^{B \times C \times H \times W}$.

Specifically, spatial priors are injected via RPE $P = R_h + R_w \in \mathbb{R}^{1 \times C \times (HW)}$, where $R_h$ and $R_w$ encode positional information along the height and width axes respectively. The position-enriched sequence is: \begin{equation} X_{\text{seq}} = X_{\text{seq}} + P^T.\end{equation} This compensates for the loss of 2D spatial structure during flattening, providing the SSM with explicit positional context prior to the causal scan.

\noindent\textbf{SSM Processing.} The enriched sequence is passed through the Mamba block $M$, resulting:\begin{equation}X_{\text{mamba}} = M(X_{\text{seq}}),\end{equation} which captures LRD across the spatial sequence. While the causal scan limits direct bidirectional interaction, the preceding RPE and the subsequent ASC module collectively recover fine-grained spatial context, as detailed in Sec.~\ref{sec:asc}. The output is reshaped back to $X_{\text{out}} \in \mathbb{R}^{B \times C \times H \times W}$.The full algorithm of the Gamba cell is provided in Appendix~A.4.

\subsection{ASC Module} 
\label{sec:asc}
\begin{figure}[ht]
    \centering
    \includegraphics[width=0.90\textwidth]
    {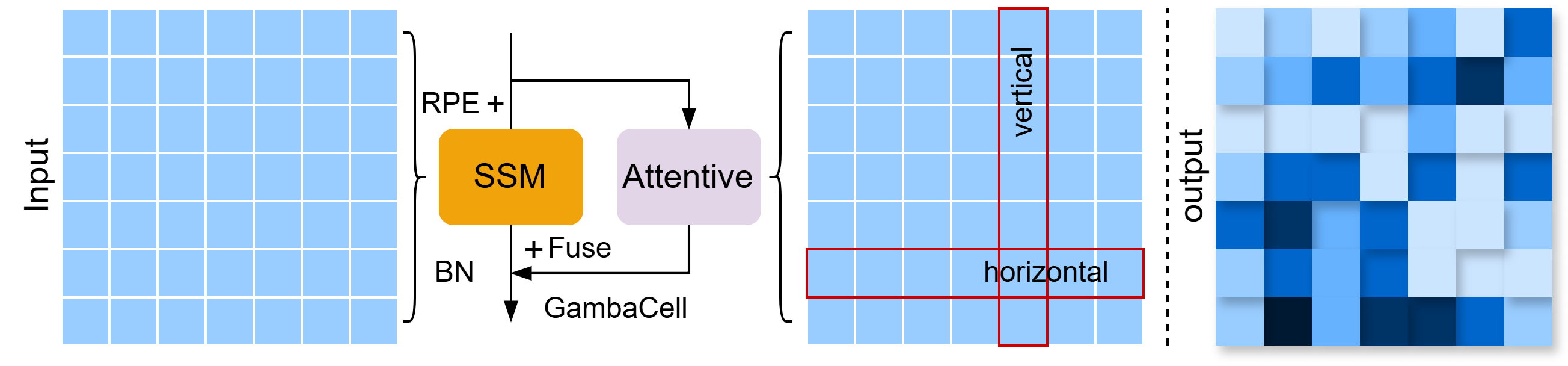}
    \caption{ASC fuses SSM and RPE to form the Gamba cell.}
    \label{fig:ASC}
\end{figure}

While SSM captures LRD through global interactions, adding 2D-RPE alone is insufficient for precise 2D spatial modeling. To address this, we introduce ASC, a lightweight attentive module that enhances fine-grained spatial context while preserving global interactions, without the need to reapply the SSM block. Fig.~\ref{fig:ASC} illustrates ASC implementation.
Unlike the original Coordinate Attention \cite{hou2021coordinate}, which encodes spatial coordinates by separately aggregating horizontal and vertical features and fusing them uniformly, we extend this design to capture directional asymmetry and channel-dependent context sensitivity. Specifically, ASC introduces per-channel adaptive gating and axis-wise learnable biases, enabling each feature channel to dynamically balance horizontal and vertical information while encoding asymmetric spatial priors.
Given an input feature $X \in \mathbb{R}^{B \times C \times H \times W}$, ASC performs coordinate pooling along height ($x_h$) and width ($x_w$), followed by shared transformations and gated fusion:
\begin{align}
f_h &= \text{Conv}_h(x_h) + b_h, \\
f_w &= \text{Conv}_w(x_w) + b_w, \\
\text{ASC}(X) &= X \odot \sigma \big( \alpha f_h + (1-\alpha) f_w \big),
\end{align}
where $\alpha \in \mathbb{R}^{1 \times C \times 1 \times 1}$ is the learnable per-channel gate, $b_h$ and $b_w$ are axis-wise learnable biases, and $\odot$ denotes element-wise multiplication.

These extensions allow ASC to capture fine-grained and asymmetric spatial dependencies along both axes efficiently, improving localized context modeling without the overhead of repeated SSM blocks.

\subsubsection{Memory and IO Efficiency}
To capture LRD, BoTNet~\cite{srinivas2021bottleneck} replaces the $3 \times 3$ convolution in ResNet-50 bottlenecks with MHSA. However, due to the quadratic memory complexity $\mathcal{O}(M^2)$ of MHSA with respect to sequence length $M$, it is restricted to the final stage ($7 \times 7$ resolution) to avoid memory explosion.

In contrast, our Gamba bottleneck integrates a Mamba-style selective SSM, whose computational complexity scales linearly with sequence length $\mathcal{O}(M)$. Moreover, we adopt the IO-aware selective scan implementation of Mamba, which reduces memory IO from $\mathcal{O}(BMEN)$ to $\mathcal{O}(BME + EN)$ by performing state updates entirely in SRAM and writing back only the output activations. Since the Gamba cell is introduced only in the fourth low-resolution stage, where $M = \frac{HW}{32^2}$, global modeling is achieved with controlled memory and computational overhead, preserving the efficiency of the original CNN backbone, with empirical validation of this scaling behavior provided in Sec.~\ref{sec:scaling}.

\subsubsection{Computation Efficiency}
The Gamba block leverages SSM to capture LRD efficiently, similar to SA in Transformers. 
Given a visual sequence $T \in \mathbb{R}^{1 \times M \times D}$ with default $E = 2D$, the computational complexities are:
\begin{align}
\Omega(\text{self-attention}) &= 4 M D^2 + 2 M^2 D, \\
\Omega(\text{SSM}) &= 3 M (2D) N + M (2D) N, \\
\Omega(\text{CNN}) &= M \cdot K^2 \cdot D^2.
\end{align}
Here, SA scales quadratically with sequence length $M$, whereas SSM scales linearly, enabling efficient processing of long sequences. Unlike CNN that capture local context efficiently, but capturing global interactions requires deeper stacks or larger RF, increasing memory and computation compared to SSM-based approaches like vGamba.

\subsection{Architectural details}
The first three stages capture local details and perform low-level feature extraction, where the input \( x \in \mathbb{R}^{C \times H \times W} \) is processed through a stem layer and standard ResNet bottleneck blocks \cite{he2016deep}. These layers extract hierarchical features while progressively reducing spatial dimensions:
\[
x_1 \in \mathbb{R}^{C_1 \times \frac{H}{4} \times \frac{W}{4}}, \quad x_2 \in \mathbb{R}^{C_2 \times \frac{H}{8} \times \frac{W}{8}}, \quad x_3 \in \mathbb{R}^{C_3 \times \frac{H}{16} \times \frac{W}{16}}
\]
In stage 4, we replace the $3\times3$ convolutional block with Gamba cell, enhancing LRD and global context modeling. The hybrid vGamba captures feature representations at deeper levels:
\[
\quad x_4 \in \mathbb{R}^{C_4 \times \frac{H}{32} \times \frac{W}{32}}
\]
Finally, global average pooling (GAP) and a fully connected layer (FC) produce the final classification output:
\[
y = \text{FC}(\text{GAP}(x_4))
\]
More architectural details of vGamba are provided in Appendix~A.4.

\newpage


\section{Experiments} \label{sec:experiments}

\begin{wraptable}{r}{0.60\textwidth}
  \centering
  \vspace{-50pt}
  \caption{Comparison on ImageNet-1K.}
  \label{tab:backbone_performance}
  \begin{tabular}{@{}l@{ }c@{ }c@{ }c@{ }c@{}}
  \toprule
    \multicolumn{5}{c}{\textbf{IMAGENET-1K}} \\
    Method & Size & FLOPs & Param & Top-1 \\
    & & (G) & (M) & (\%) \\
    \midrule
    ResNet-50 \cite{he2016deep} & $224^2$ & 3.9 & 25  & 76.2 \\
    ResNet-101 \cite{he2016deep} & $224^2$ & 7.6 & 45 & 77.4 \\
    ResNet-152 \cite{he2016deep} & $224^2$ & 11.3 & 60 & 78.3 \\
    BoT50 \cite{srinivas2021bottleneck} & $224^2$ & 4.1 & 20.8 & 78.3 \\
    EffNet-B2 \cite{tan2019efficientnet} & $224^2$ & 1.0 & 9 & 80.1  \\
    EffNet-B3 \cite{tan2019efficientnet} & $300^2$ & 1.8 & 12 & 81.6 \\
    RegNetY-4GF \cite{radosavovic2020designing} & $224^2$ & 4.0 & 21 & 80.0 \\
    RegNetY-8GF \cite{radosavovic2020designing} & $224^2$ & 8.0 & 39 & 81.7 \\
    \midrule
    ViT-B/16 \cite{dosovitskiy2021image} & $384^2$ & 55.4 & 86 & 77.9 \\
    ViT-L/16 \cite{dosovitskiy2021image} & $384^2$ & 190.7 & 307 & 76.5 \\
    DeiT-S \cite{touvron2021training} & $224^2$ & 4.6 & 22 & 79.8 \\
    DeiT-B \cite{touvron2021training} & $224^2$ & 17.5 & 86 & 81.8 \\
    Swin-T \cite{liu2021swin} & $224^2$ & 4.5 & 29 & 81.3 \\
    CoaT-Lite-S \cite{dai2021coatnet}& $224^2$ & 4.1 & 19.8 & 82.3 \\
    CrossViT-B \cite{chen2021crossvit} & $240^2$ & 20.1 & 105 & 82.2 \\
    \midrule
    Vim-S \cite{zhu2024vision} & $224^2$ & 5.3 & 26 & 80.3 \\
    Vim-B \cite{zhu2024vision} & $224^2$ & -- & 98 & 81.9 \\
    EffVMamba-T \cite{pei2024efficientvmamba} & $224^2$ & 0.8 & 6 & 76.5 \\
    EffVMamba-S \cite{pei2024efficientvmamba} & $224^2$ & 1.3 & 11 & 78.7 \\
    S4ND-ViT-B \cite{nguyen2022s4nd} & $224^2$ & 17.1 & 89 & 80.4 \\
    VMamba-T \cite{liu2025vmamba} & $224^2$ & 4.9 & 30 & 82.6 \\
    MambaVision-T \cite{hatamizadeh2024mambavision} & $224^2$ & 4.4 & 31.8 & 82.3 \\
    \midrule
    vGamba-B & $224^2$ & 3.8 & 19.4 & 81.1 \\
    vGamba-L & $224^2$ & 7.6 & 38.4 & 82.8 \\
    vGamba-X & $224^2$ & 11.3 & 54 & 83.2 \\
    \bottomrule
  \end{tabular}
  \vspace{-40pt}
\end{wraptable}

We evaluated vGamba on ImageNet-1K for classification (Section \ref{sec:classification}), ADE20K for segmentation (Section \ref{sec:segmentation}), and COCO for detection (Section \ref {sec:detection}). Then qualitative experiments on the AID dataset (Section \ref{sec:aid}) and performed ablation studies (Section \ref{sec:ablation}). Experiments are conducted on NVIDIA GeForce RTX 4090 GPU, featuring 24 GB of GDDR6X VRAM, CUDA Compute Capability 8.9, and driver version 570.144 with CUDA 12.8 support.

\subsection{Classification}\label{sec:classification}
\textbf{Settings.} The ImageNet-1K dataset \cite{deng2009imagenet} contains 1.28M training images and 50K validation images across 1,000 categories. Training follows ConvNeXt \cite{liu2022convnet} settings, using augmentations like random cropping, flipping, label-smoothing, mixup, and random erasing. For $224^2$ input images, we optimize with AdamW with momentum 0.9, batch size 64, and weight decay 0.05. Then trained vGamba models for 250 epochs with a cosine schedule and EMA, starting with a $1\times 10^{-3}$ learning rate.

\textbf{Results.} Table~\ref{tab:backbone_performance} reports ImageNet-1K classification results. Among the three bottleneck designs, vGamba-B outperforms ResNet-50 by $+4.9\%$ and BoTNet by $+2.8\%$ in Top-1 accuracy, while reducing FLOPs by $0.1$G and $0.3$G and parameters by $5.6$M and $1.4$M respectively, demonstrating a better efficiency--accuracy trade-off among bottleneck designs. Among Transformer-based models, vGamba-B surpasses DeiT-S by $+1.3\%$ with $0.9$G fewer FLOPs and $3.1$M fewer parameters. Among SSM-based models, vGamba-B exceeds ViM-S by $+0.8\%$ with $1.6$G lower compute. VMamba-T achieves $+1.5\%$ higher accuracy than vGamba-B, however it requires $1.2$G more FLOPs and $11.1$M more parameters, confirming that vGamba-B achieves competitive accuracy without the overhead of multi-directional SSM replication.

\subsection{Segmentation}\label{sec:segmentation}

\begin{wraptable}{r}{0.59\textwidth}
  \centering
      \vspace{-35pt}
  \caption{Comparison of semantic segmentation results on ADE20K using UperNet. FLOPs are measured with an input size of 512\texttimes2048.}
  \label{tab:segmentation_results}
  \begin{tabular}{l@{ }c@{ }c@{ }c@{ }c@{}}
    \toprule
    \multicolumn{5}{c}{\textbf{ADE20K}} \\
    Backbone & Param & FLOP & mIoU & mIoU \\
     & (M) & (G) & (SS) & (MS) \\
    \midrule
    ResNet-50 \cite{he2016deep} & 67 & 953 & 42.1 & 42.8 \\
    ResNet-101 \cite{he2016deep} & 85 & 1030 & 42.9 & 44.0 \\
    ConvNeXt-T \cite{liu2022convnet} & 60 & 939 & 46.0 & 46.7 \\
    ConvNeXt-S \cite{liu2022convnet} & 82 & 1027 & 48.7 & 49.6 \\
    \midrule
    DeiT-S + MLN \cite{touvron2021training} & 58 & 1217 & 43.8 & 45.1 \\
    DeiT-B + MLN \cite{touvron2021training} & 144 & 2007 & 45.5 & 47.2 \\
    Swin-T \cite{liu2021swin} & 60 & 945 & 44.5 & 45.8 \\
    Swin-S \cite{liu2021swin} & 81 & 1039 & 47.6 & 49.5 \\
    Swin-B \cite{liu2021swin} & 121 & 1188 & 48.1 & - \\
    
    \midrule
    Vim-S \cite{zhu2024vision} & 46 & -- & 44.9 & -- \\
    VMamba-T \cite{liu2025vmamba} & 62 & 949 & 47.9 & 48.8 \\
    VMamba-S \cite{liu2025vmamba} & 82 & 1028 & 50.6 & 51.2 \\
    VMamba-B \cite{liu2025vmamba} & 122 & 1170 & 51.0 & 51.6 \\
    Eff.VMamba-T \cite{pei2024efficientvmamba} & 14 & 230 & 38.9 & 39.3 \\
    Eff.VMamba-S \cite{pei2024efficientvmamba} & 29 & 505 & 41.5 & 42.1 \\
    Eff.VMamba-B \cite{pei2024efficientvmamba} & 65 & 930 & 46.5 & 47.3 \\
    \midrule
    vGamba-B & 55 & 941 & 46.6 & 46.7 \\
    vGamba-L & 71 & 1019 & 46.9 & 47.8 \\
    \bottomrule
  \end{tabular}
    \vspace{-20pt}
\end{wraptable}

\textbf{Settings.} We conducted semantic segmentation experiments using ADE20K dataset \cite{zhou2019semantic} within the UperNet framework \cite{xiao2018unified}. The backbone network was initialized with pre-trained weights from ImageNet-1K \cite{deng2009imagenet}, while the remaining components were randomly initialized. Model optimization was performed using the AdamW optimizer with a batch size of 16. The model was trained for 160k iterations to ensure thorough learning. 

\textbf{Results.} Table~\ref{tab:segmentation_results} reports semantic segmentation results on ADE20K using UperNet. Among CNN backbones, vGamba-B improves over ResNet-50 by $+4.5\%$ SS mIoU with $14$M fewer parameters and comparable FLOPs, confirming that the Gamba cell enhances spatial context modeling over pure CNN feature extraction. Compared to Transformer-based backbones, vGamba-B surpasses DeiT-S+MLN by $+2.8\%$ SS mIoU while requiring $276$G fewer FLOPs, and matches Swin-T ($+2.1\%$) at comparable parameters and compute. Among SSM-based models, vGamba-B achieves $+1.7\%$ over ViM-S and comparable performance to Eff.VMamba-B ($+0.1\%$) with similar compute. VMamba-T and VMamba-S achieve higher mIoU than vGamba-B by $1.3\%$ and $4.0\%$ respectively, however both require additional parameters and FLOPs, reflecting the cost of multi-directional SSM replication. Overall, vGamba-B achieves a competitive efficiency--accuracy trade-off across all backbone categories.

\subsection{Object Detection}\label{sec:detection}
\textbf{Settings.} We evaluated our model on object detection and instance segmentation tasks using the COCO 2017 dataset \cite{Lin2014MicrosoftCOCO}, which consists of approximately 118K training images and 5K validation images. vGamba was utilized as the backbone and integrated into the Mask R-CNN framework \cite{he2017mask} for feature extraction. The model was initialized with weights pre-trained on ImageNet-1K (300 epochs) and trained for 12 epochs (1×) and 36 epochs (3×).

\begin{wraptable}{r}{0.6\textwidth}
  \centering
  \vspace{-12pt}
  \caption{Cascade Mask R-CNN detection results.}
  \label{tab:mask_rcnn_results}
  \begin{tabular}{lcccc}
    \toprule
    Backbone & FLOPs & Params & AP$_b$ & AP$_m$ \\
    & (G) & (M) & & \\
    \midrule
    \multicolumn{5}{c}{\textbf{Mask R-CNN 1× schedule}} \\
    Swin-T \cite{liu2021swin} & 267G & 48M & 42.7 & 39.3 \\
    ConvNeXt-T \cite{liu2022convnet} & 262G & 48 & 44.2 & 40.1 \\
    VMamba-T \cite{liu2025vmamba} & 271G & 50 & 47.3 & 42.7 \\
    Swin-S \cite{liu2021swin} & 354G & 69 & 44.8 & 40.9 \\
    ConvNeXt-S \cite{liu2022convnet} & 348G & 70 & 45.4 & 41.8 \\
    VMamba-S \cite{liu2025vmamba} & 349G & 70 & 48.7 & 43.7 \\
    Swin-B \cite{liu2021swin} & 496G & 107 & 46.9 & 42.3 \\
    ConvNeXt-B \cite{liu2022convnet} & 486G & 108 & 47.0 & 42.7 \\
    VMamba-B \cite{liu2025vmamba} & 485G & 108 & 49.2 & 44.1 \\
    vGamba-B & 225G & 49 & 44.9 & 42.0 \\
    vGamba-L & 254G & 71 & 49.8& 45.3 \\
    
    \midrule
    \multicolumn{5}{c}{\textbf{Mask R-CNN 3× MS schedule}} \\
    Swin-T \cite{liu2021swin} & 267G & 48 & 46.0 & 41.6 \\
    ConvNeXt-T \cite{liu2022convnet} & 262G & 48 & 46.2 & 41.7 \\
    VMamba-T \cite{liu2025vmamba} & 271G & 50 & 48.8 & 43.7 \\
    Swin-S \cite{liu2021swin} & 354G & 69 & 48.2 & 43.2 \\
    ConvNeXt-S \cite{liu2022convnet} & 348G & 70 & 47.9 & 42.9 \\
    VMamba-S \cite{liu2025vmamba} & 349G & 70 & 49.9 & 44.2 \\
    vGamba-L & 254G & 71 & 51.3 & 46.1 \\
    
    \midrule
    \multicolumn{5}{c}{\textbf{Cascade Mask R-CNN 3× MS schedule}} \\
    DeiT-Small/16 \cite{touvron2021training} & 889G & 80 & 48.0 & 41.4 \\
    ResNet-50 \cite{he2016deep} & 739G & 82 & 46.3 & 40.1 \\
    Swin-T \cite{liu2021swin} & 745G & 86 & 50.4 & 43.7 \\
    ConvNeXt-T \cite{liu2022convnet} & 741G & 86 & 50.4 & 43.7 \\
    MambaVision-T \cite{hatamizadeh2024mambavision} & 740G & 86 & 51.0 & 44.3 \\
    Swin-S \cite{liu2021swin} & 838G & 107 & 51.9 & 45.0 \\
    ConvNeXt-S \cite{liu2022convnet} & 827G & 108 & 51.9 & 45.0 \\
    MambaVision-S \cite{hatamizadeh2024mambavision} & 828G & 108 & 52.1 & 45.2 \\
    Swin-B \cite{liu2021swin} & 982G & 145 & 51.9 & 45.0 \\
    ConvNeXt-B \cite{liu2022convnet} & 964G & 146 & 52.7 & 45.6 \\
    MambaVision-B \cite{hatamizadeh2024mambavision} & 964G & 145 & 52.8 & 45.7 \\
    vGamba-B & 727G & 76 & 50.8 & 43.7 \\
    vGamba-L & 905G & 123 & 52.0 & 45.3 \\
    \bottomrule
      \vspace{-50pt}
  \end{tabular}
\end{wraptable}

\textbf{Results.} Table~\ref{tab:mask_rcnn_results} reports object 
detection and instance segmentation results across three schedules.

\noindent\textbf{Mask R-CNN 1×.} vGamba-B achieves comparable 
detection to VMamba-T ($-2.4$ AP$_b$) while requiring $46$G fewer 
FLOPs at similar parameter count, and surpasses both Swin-T and 
ConvNeXt-T by $+2.2$ and $+0.7$ AP$_b$ respectively with fewer 
FLOPs, reflecting the efficiency advantage of the single-block 
SSM design.

\noindent\textbf{Mask R-CNN 3× MS.} vGamba-L outperforms VMamba-S 
by $+1.4$ AP$_b$ and $+1.9$ AP$_m$ with $95$G fewer FLOPs and 
comparable parameters, confirming that vGamba benefits 
proportionally more from longer training schedules relative to 
its compute cost.

\noindent\textbf{Cascade Mask R-CNN 3× MS.} vGamba-B surpasses 
Swin-T and ConvNeXt-T by $+0.4$ AP$_b$ with $10$M fewer parameters 
at comparable FLOPs. MambaVision-T achieves $+0.2$ AP$_b$ higher 
than vGamba-B with $10$M more parameters, reflecting a competitive 
efficiency--accuracy trade-off among SSM-based detection backbones.

\subsection{Scaling Performance}
\label{sec:scaling}
We evaluate the scaling performance of ResNet-50, BoTNet, and vGamba backbones across input resolutions of $512$, $640$, $768$, $1024$, and $2048$. For each setting, we measure inference throughput (FPS) and peak GPU memory consumption. This analysis provides insight into each model’s computational efficiency and practical suitability under varying resolution and hardware constraints.
\begin{figure}[htbp]
    \centering
    \begin{tabular}{cc}
        \includegraphics[width=0.49\linewidth]{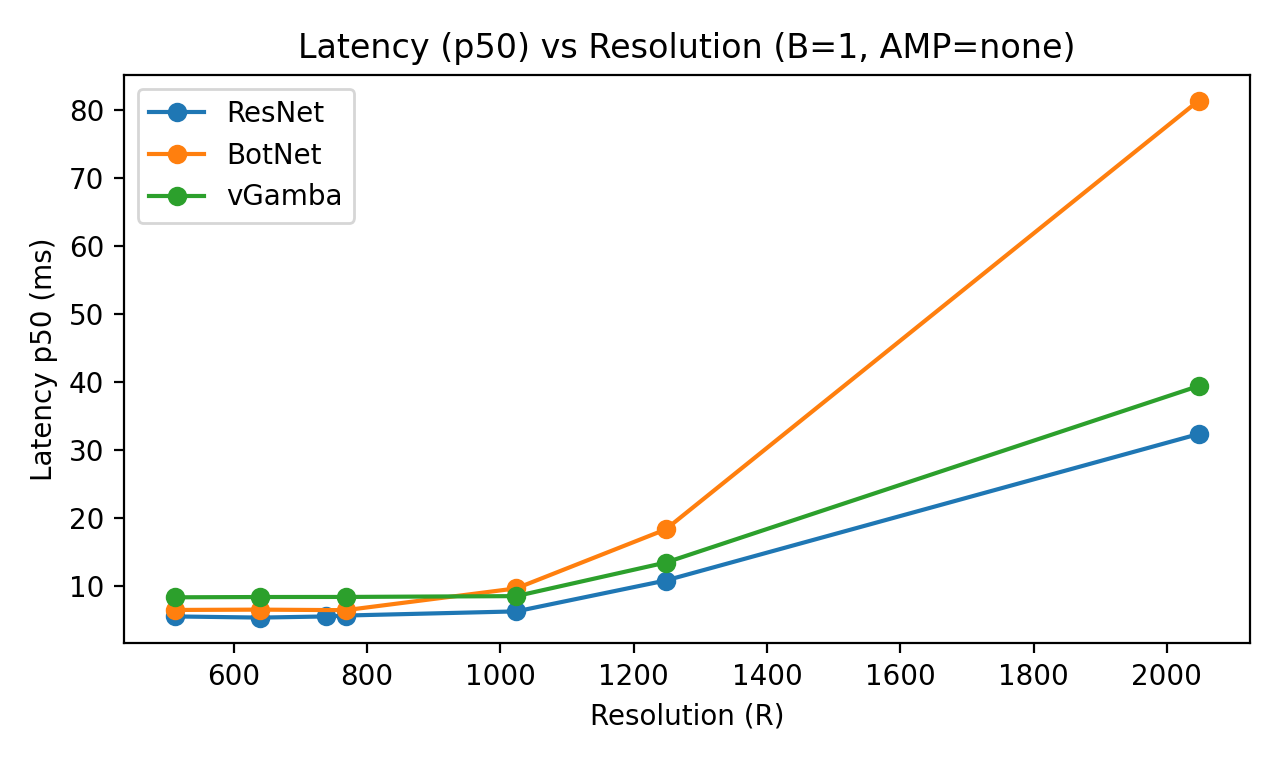} &
        \includegraphics[width=0.49\linewidth]{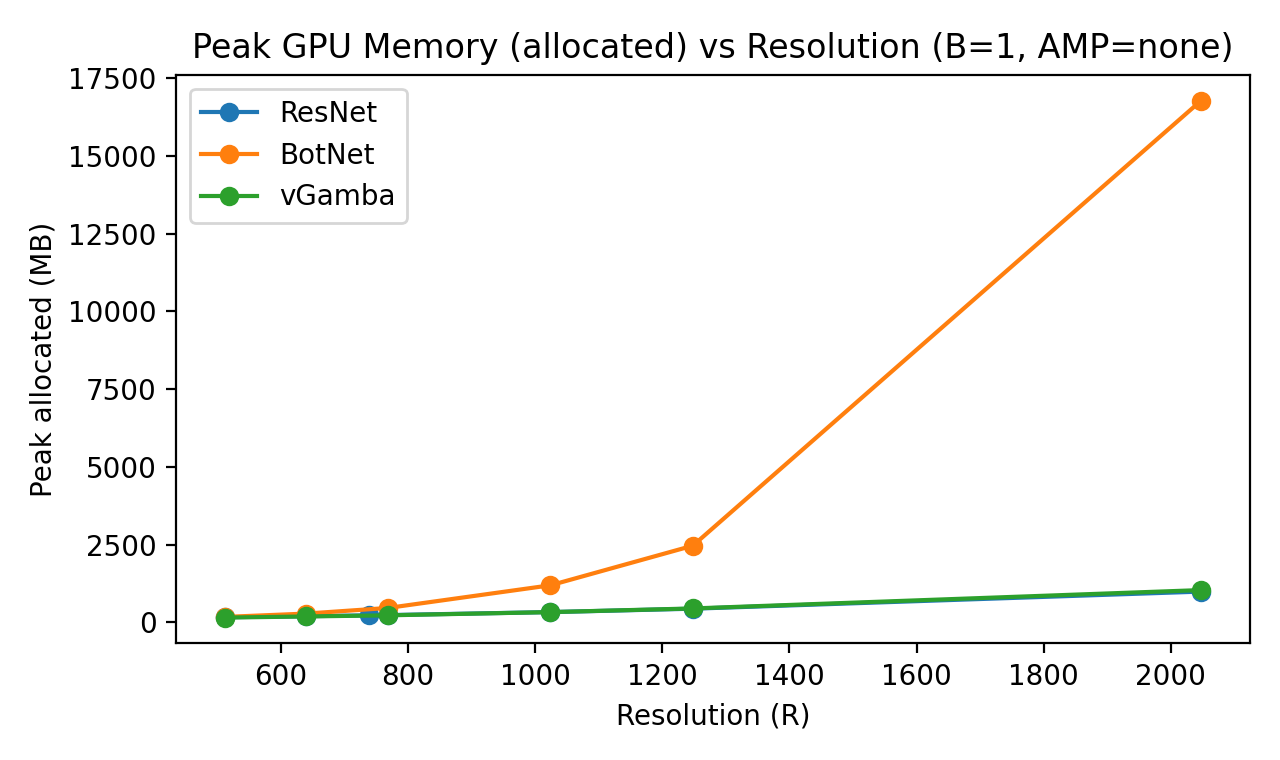} \\
        (a) FPS comparison (speed) & (b) Memory usage
    \end{tabular}
    \caption{Scaling behavior of ResNet-50, BoTNet and vGamba across increasing input resolutions. (a) Throughput decreases more rapidly for BoTNet as spatial size grows. (b) Peak memory usage shows superlinear growth for BoTNet, while vGamba follows a near-linear trend comparable to ResNet-50.}
    \label{fig:scaling_performance}
\end{figure}
As shown in Figure ~\ref{fig:scaling_performance}, vGamba achieves computational and memory scaling behavior comparable to ResNet-50~\cite{he2016deep}, indicating near-CNN-level efficiency. In contrast, BoTNet~\cite{srinivas2021bottleneck} demonstrates substantially higher memory growth and reduced throughput as resolution increases. For example, at $2048 \times 2048$ resolution, vGamba is $2.07\times$ faster than BoTNet (39.41 ms vs.\ 81.47 ms) and reduces peak GPU memory usage by $93.8\%$ (1.03 GB vs.\ 16.78 GB) under identical inference settings. Additional details are provided in Appendix A.1.

\subsection{Qualitative Experiments} \label{sec:aid}
To understand model behavior beyond quantitative metrics, we conduct qualitative analyses focusing on interpretability and receptive field characteristics. Specifically, we investigate (1) attention localization using Hi-Res-CAM \cite{draelos2020use} and (2) effective receptive field (ERF)~\cite{luo2016understanding} distributions across layers. These analyses reveal how the proposed architecture captures long-range dependencies and contextual information in complex aerial scenes.

\subsubsection{Hi-Res-CAM} 
\begin{figure*}[htbp]
    \centering
    \includegraphics[width=.93\textwidth]{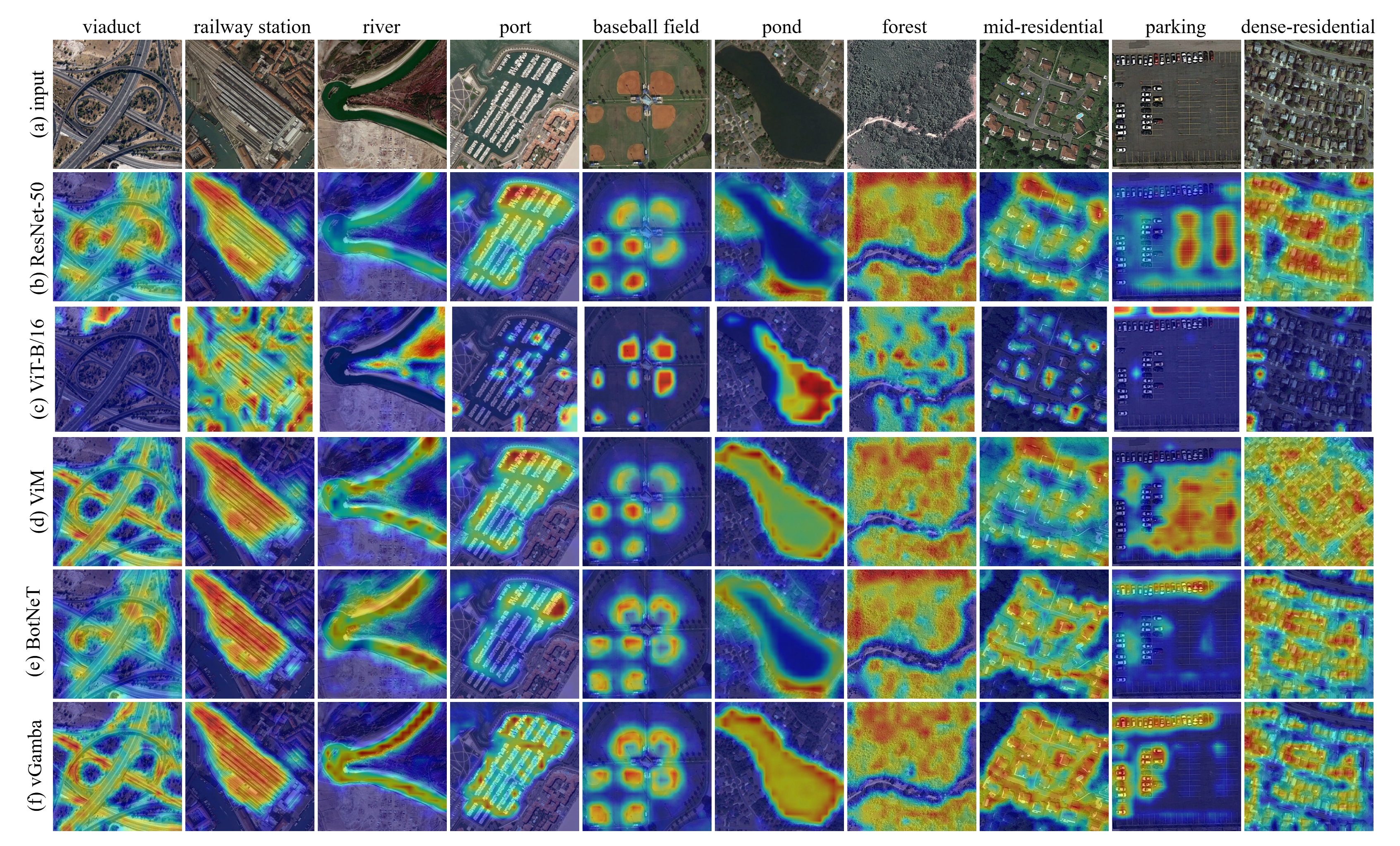}
    \caption{Hi-Res-CAM visualization on AID, highlighting model decision regions.}
\label{fig:aidd}
\end{figure*}

was applied on the AID dataset \cite{xia2017aid} (see Appendix A.2 for implementation details) to visualize attention regions. AID contains large-scale aerial scenes with long-range spatial dependencies. As shown in Fig.~\ref{fig:aidd}, heatmaps indicate that the model attends to both discriminative local textures and broader semantic layouts, such as runways, roads, and surrounding infrastructure. Unlike CNNs such as ResNet~\cite{he2016deep}, which focus on isolated local patches, ViT-B/16~\cite{dosovitskiy2021image} captures long-range dependencies, though interpretability can be affected by high norms during inference \cite{darcet2023vision} and computation cost is higher. ViM\cite{zhu2024vision} and BotNet~\cite{srinivas2021bottleneck} also performs competitively. Our approach captures globally coherent regions aligned with scene semantics, confirming that it integrates local and long-range cues and produces meaningful decision regions.

\subsubsection{ERF} 
\begin{figure*}[htbp]
    \centering
    \begin{tabular}{@{ }c c@{ }}
        \begin{minipage}[c]{0.05\textwidth}
            \centering
            \rotatebox{90}{ResNet-50}
        \end{minipage} &
        \begin{minipage}[c]{0.85\textwidth}
            \includegraphics[width=\textwidth]{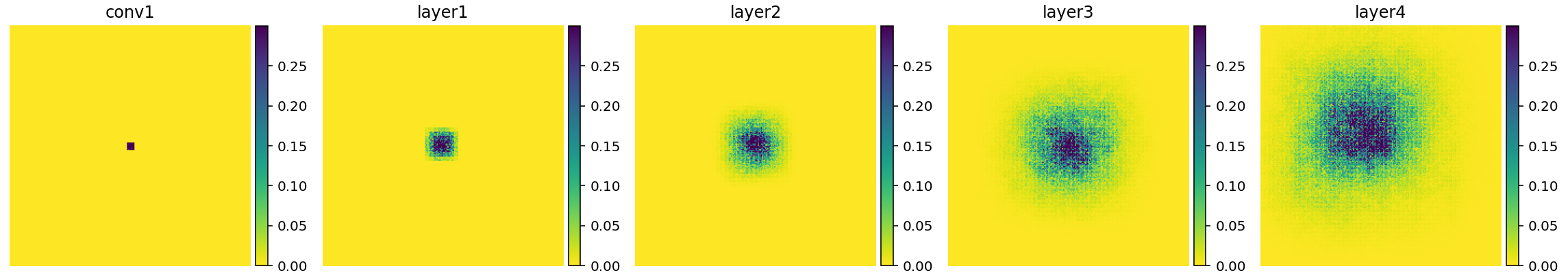}
        \end{minipage} \\
        
        \begin{minipage}[c]{0.05\textwidth}
            \centering
            \rotatebox{90}{BoTNet}
        \end{minipage} &
        \begin{minipage}[c]{0.85\textwidth}
            \includegraphics[width=\textwidth]{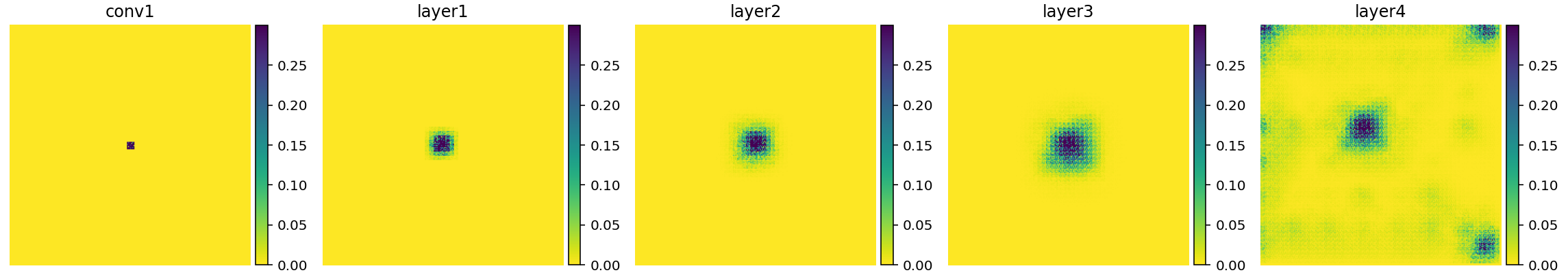}
        \end{minipage} \\
        
        \begin{minipage}[c]{0.05\textwidth}
            \centering
            \rotatebox{90}{vGamba}
        \end{minipage} &
        \begin{minipage}[c]{0.85\textwidth}
            \includegraphics[width=\textwidth]{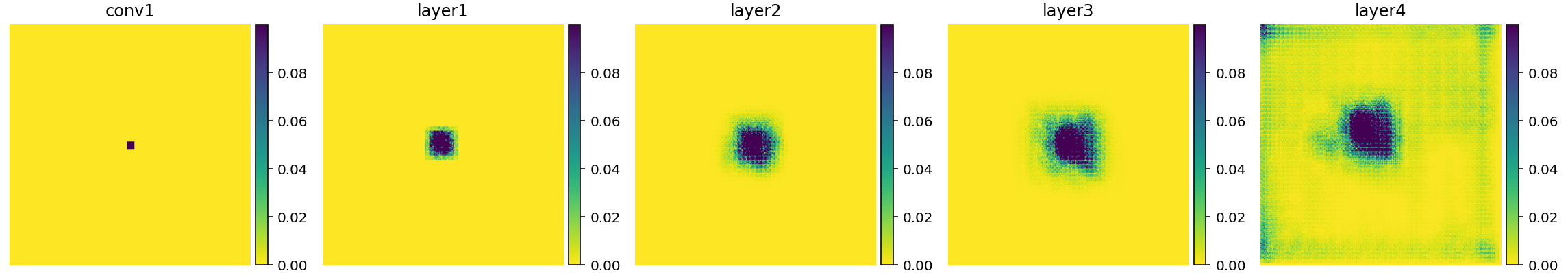}
        \end{minipage} \\
    \end{tabular}
    \caption{ERF comparison among ResNet-50, BoTNet, and vGamba across layers.}
    \label{fig:erf}
\end{figure*}

quantifies how much of the input image effectively contributes to an output activation, revealing how models capture LRD beyond their theoretical receptive field~\cite{luo2016understanding}. We compare ResNet-50, BoTNet, and vGamba under identical settings (see Appendix A.2 for implementation details). As shown in Fig.~\ref{fig:erf}, ResNet-50 exhibits local ERFs, while BoTNet achieves global coverage via attention and vGamba attains comparable global context more efficiently through its SSM-based design. ERFs remain local in layers 0--3 but expand globally in layer 4, validating the Gamba cell placement for efficient LRD modeling.

\newpage

\subsection{Ablation}
\label{sec:ablation}
\noindent\textbf{Effects of Components.}Table~\ref{tab:ablation_combined} (a) ablates each GCell component. Adding RPE alone improves top-1 to $+2.2\%$ gain without incuring computational complexity, confirming that positional priors compensate for spatial structure lost during flattening. Adding ASC alone results in a stronger $+2.6\%$ gain at a marginal cost of $+0.5$G FLOPs. Combining both achieves the best result at $+2.8\%$, showing that RPE and ASC are complementary rather than redundant.

\noindent\textbf{Effects of Attention.} ~Table~\ref{tab:ablation_combined} (b) compares ASC against established attention alternatives under identical settings. SE Block~\cite{hu2018squeeze} and CBAM~\cite{woo2018cbam} underperform by $-1.1\%$ and $-0.8\%$ respectively, SE operates channel-wise only, while CBAM lacks directional bias. Coordinate Attention~\cite{hou2021coordinate} falls short by $-0.5\%$, lacking per-channel gating and axis-wise biases for asymmetric spatial priors. ASC achieves the highest accuracy at comparable FLOPs and parameters, confirming that per-channel gating and axis-wise biases are the key differentiators. See Appendix A.3 for additional ablations.

\begin{table}[h]
  \centering
  \caption{Ablation (a) Gamba cell components and 
  (b) spatial attention mechanisms.}
  \label{tab:ablation_combined}
  \begin{minipage}{0.48\textwidth}
    \centering
    \begin{tabular}{@{}l@{ }c@{ }c@{ }c@{}}
      \toprule
      \textbf{(a) Components} & \textbf{Top-1} & \textbf{FLOPs} & \textbf{Param} \\
      \midrule
      None              & 78.3\% & 3.38G & 19.38M \\
      RPE~\textit{only} & 80.5\% & 3.38G & 19.38M \\
      ASC~\textit{only} & 80.9\% & 3.88G & 19.45M \\
      RPE + ASC         & 81.1\% & 3.88G & 19.45M \\
      \bottomrule
    \end{tabular}
  \end{minipage}
  \hfill
  \begin{minipage}{0.48\textwidth}
    \centering
    \begin{tabular}{@{}l@{ }c@{ }c@{ }c@{}}
      \toprule
      \textbf{(b) Attention} & \textbf{Top-1} & \textbf{FLOPs} & \textbf{Param} \\
      \midrule
      SE Block \cite{hu2018squeeze}              & 80.0\% & 3.87G & 19.47M \\
      CBAM \cite{woo2018cbam}                    & 80.3\% & 3.87G & 19.47M \\
      Coord. Attn. \cite{hou2021coordinate}      & 80.6\% & 3.88G & 19.45M \\
      ASC (Ours)                                 & 81.1\% & 3.88G & 19.45M \\
      \bottomrule
    \end{tabular}
  \end{minipage}
\end{table}

\section{Conclusion}
This work advances bottleneck design for LRD modeling by demonstrating that a single, lightweight SSM-enhanced bottleneck can replace both CNN and Transformer bottlenecks while efficiently capturing global context. The Gamba bottleneck embeds a spatially aware SSM (the Gamba cell) and the ASC module into a compact structure that maintains representational strength. The resulting backbone, vGamba, outperforms CNN- and Transformer-based alternatives across downstream tasks, capturing LRD more accurately and with smoother convergence while using lower memory and FLOPs. Overall, vGamba provides a practical, scalable alternative to existing visual backbones.

\subsubsection{Limitations.} While vGamba demonstrates strong performance across standard benchmarks, its effectiveness on domain-specific tasks requiring extreme precision, such as medical imaging, remains unevaluated. Furthermore, restricting the Gamba cell to stage 4 only may limit fine-grained feature modulation in earlier stages, which could affect performance on dense prediction tasks such as segmentation, as reflected in the segmentation results (Sec.~\ref{sec:segmentation}).


\bibliographystyle{splncs04}
\bibliography{main}

@String(CVPR  = {IEEE Conf. Comput. Vis. Pattern Recog.})

@String(ICCV  = {Int. Conf. Comput. Vis.})

@String(ECCV  = {Eur. Conf. Comput. Vis.})

@String(NeurIPS = {Adv. Neural Inform. Process. Syst.})

@String(ICLR  = {Int. Conf. Learn. Represent.})

@String(CVPR  = {CVPR})

@String(ICCV  = {ICCV})

@String(ECCV  = {ECCV})

@String(NeurIPS = {NeurIPS})

@String(ICLR  = {ICLR})

@inproceedings{szegedy2015going,
  author = {Szegedy, Christian and Liu, Wei and Jia, Yangqing and Sermanet, Pierre and Reed, Scott and Anguelov, Dragomir and Erhan, Dumitru and Vanhoucke, Vincent and Rabinovich, Andrew},
  title = {Going deeper with convolutions},
  booktitle = {Proceedings of the IEEE conference on computer vision and pattern recognition},
  pages = {1--9},
  year = {2015}
}

@inproceedings{he2016deep,
  author = {He, Kaiming and Zhang, Xiangyu and Ren, Shaoqing and Sun, Jian},
  title = {Deep residual learning for image recognition},
  booktitle = {Proceedings of the IEEE conference on computer vision and pattern recognition},
  pages = {770--778},
  year = {2016}
}

@inproceedings{chollet2017xception,
  author = {Chollet, François},
  title = {Xception: Deep learning with depthwise separable convolutions},
  booktitle = {Proceedings of the IEEE conference on computer vision and pattern recognition},
  pages = {1251--1258},
  year = {2017}
}

@inproceedings{tan2019efficientnet,
  author = {Tan, Mingxing and Le, Quoc},
  title = {Efficientnet: Rethinking model scaling for convolutional neural networks},
  booktitle = {International conference on machine learning},
  pages = {6105--6114},
  year = {2019}
}

@inproceedings{radosavovic2020designing,
  author = {Radosavovic, Ilija and Kosaraju, Raj Prateek and Girshick, Ross and He, Kaiming and Dollár, Piotr},
  title = {Designing network design spaces},
  booktitle = {Proceedings of the IEEE/CVF conference on computer vision and pattern recognition},
  pages = {10428--10436},
  year = {2020}
}

@inproceedings{ding2022scaling,
  author = {Ding, Xiaohan and Zhang, Xiangyu and Han, Jungong and Ding, Guiguang},
  title = {Scaling up your kernels to 31x31: Revisiting large kernel design in CNNs},
  booktitle = {Proceedings of the IEEE/CVF Conference on Computer Vision and Pattern Recognition (CVPR)},
  pages = {11963--11975},
  year = {2022}
}

@inproceedings{peng2017large,
  title={Large kernel matters -- Improve semantic segmentation by global convolutional network},
  author={Peng, Chao and Zhang, Xiangyu and Yu, Gang and Luo, Guiming and Sun, Jian},
  booktitle={Proceedings of the IEEE Conference on Computer Vision and Pattern Recognition (CVPR)},
  pages={4353--4361},
  year={2017}
}

@inproceedings{luo2016understanding,
  title     = {Understanding the Effective Receptive Field in Deep Convolutional Neural Networks},
  author    = {Luo, Wenjie and Li, Yujia and Urtasun, Raquel and Zemel, Richard},
  booktitle = {Advances in Neural Information Processing Systems (NeurIPS)},
  volume    = {29},
  year      = {2016}
}

@inproceedings{hou2021coordinate,
  title={Coordinate attention for efficient mobile network design},
  author={Hou, Qibin and Zhou, Daquan and Feng, Jiashi},
  booktitle={Proceedings of the IEEE/CVF Conference on Computer Vision and Pattern Recognition (CVPR)},
  pages={13713--13722},
  year={2021}
}

@inproceedings{yu2017dilated,
  author = {Yu, Fisher and Koltun, Vladlen and Funkhouser, Thomas},
  title = {Dilated residual networks},
  booktitle = {Proceedings of the IEEE conference on computer vision and pattern recognition},
  pages = {472--480},
  year = {2017}
}

@inproceedings{wang2018non,
  author = {Wang, Xiaolong and Girshick, Ross and Gupta, Abhinav and He, Kaiming},
  title = {Non-local neural networks},
  booktitle = {Proceedings of the IEEE conference on computer vision and pattern recognition},
  pages = {7794--7803},
  year = {2018}
}

@inproceedings{dosovitskiy2021image,
  author = {Dosovitskiy, Alexey and Beyer, Lucas and Kolesnikov, Alexander and Weissenborn, Dirk and Zhai, Xiaohua and Unterthiner, Thomas and Dehghani, Mostafa and Minderer, Matthias and Heigold, Georg and Gelly, Sylvain and Uszkoreit, Jakob and Houlsby, Neil},
  title = {An Image is Worth 16x16 Words: Transformers for Image Recognition at Scale},
  booktitle = {ICLR},
  year = {2021}
}

@inproceedings{yang2024efficient,
  author = {Yang, Yuedong and Chiang, Hung-Yueh and Li, Guihong and Marculescu, Diana and Marculescu, Radu},
  title = {Efficient low-rank backpropagation for vision transformer adaptation},
  journal = {Advances in Neural Information Processing Systems},
  volume = {36},
  year = {2024}
}

@inproceedings{ma2021luna,
  author = {Ma, Xuezhe and Kong, Xiang and Wang, Sinong and Zhou, Chunting and May, Jonathan and Ma, Hao and Zettlemoyer, Luke},
  title = {Luna: Linear unified nested attention},
  journal = {Advances in Neural Information Processing Systems},
  volume = {34},
  pages = {2441--2453},
  year = {2021}
}

@article{song2021ufo,
  author = {Song, Jeong-geun},
  title = {Ufo-vit: High performance linear vision transformer without softmax},
  journal = {arXiv preprint arXiv:2109.14382},
  year = {2021}
}

@inproceedings{chen2023sparsevit,
  author = {Chen, Xuanyao and Liu, Zhijian and Tang, Haotian and Yi, Li and Zhao, Hang and Han, Song},
  title = {Sparsevit: Revisiting activation sparsity for efficient high-resolution vision transformer},
  booktitle = {Proceedings of the IEEE/CVF Conference on Computer Vision and Pattern Recognition},
  pages = {2061--2070},
  year = {2023}
}

@article{gu2023mamba,
  author = {Gu, Albert and Dao, Tri},
  title = {Mamba: Linear-time sequence modeling with selective state spaces},
  journal = {arXiv preprint arXiv:2312.00752},
  year = {2023}
}

@article{zhu2024vision,
  author = {Zhu, Lianghui and Liao, Bencheng and Zhang, Qian and Wang, Xinlong and Liu, Wenyu and Wang, Xinggang},
  title = {Vision Mamba: Efficient Visual Representation Learning with Bidirectional State Space Model},
  journal = {arXiv},
  year = {2024},
  url = {https://arxiv.org/abs/2401.09417}
}

@article{liu2025vmamba,
  author = {Liu, Yue and Tian, Yunjie and Zhao, Yuzhong and Yu, Hongtian and Xie, Lingxi and Wang, Yaowei and Ye, Qixiang and Jiao, Jianbin and Liu, Yunfan},
  title = {Vmamba: Visual state space model},
  journal = {Advances in neural information processing systems},
  volume = {37},
  pages = {103031--103063},
  year = {2025}
}

@article{hatamizadeh2024mambavision,
  author = {Hatamizadeh, Ali and Kautz, Jan},
  title = {Mambavision: A hybrid mamba-transformer vision backbone},
  journal = {arXiv preprint arXiv:2407.08083},
  year = {2024}
}

@inproceedings{srinivas2021bottleneck,
  author    = {Aravind Srinivas and Tsung-Yi Lin and Niki Parmar and Jonathon Shlens and Pieter Abbeel and Ashish Vaswani},
  title     = {Bottleneck Transformers for Visual Recognition},
  booktitle = {Proceedings of the IEEE/CVF Conference on Computer Vision and Pattern Recognition},
  pages     = {16519--16529},
  year      = {2021}
}

@inproceedings{deng2009imagenet,
  title={Imagenet: A large-scale hierarchical image database},
  author={Deng, Jia and Dong, Wei and Socher, Richard and Li, Li-Jia and Li, Kai and Fei-Fei, Li},
  booktitle={2009 IEEE conference on computer vision and pattern recognition},
  pages={248--255},
  year={2009},
  organization={Ieee}
}

@inproceedings{liu2022convnet,
  title={A convnet for the 2020s},
  author={Liu, Zhuang and Mao, Hanzi and Wu, Chao-Yuan and Feichtenhofer, Christoph and Darrell, Trevor and Xie, Saining},
  booktitle={Proceedings of the IEEE/CVF Conference on Computer Vision and Pattern Recognition},
  pages={11976--11986},
  year={2022}
}

@inproceedings{xiao2018unified,
  author    = {Tete Xiao and Yingcheng Liu and Bolei Zhou and Yuning Jiang and Jian Sun},
  title     = {Unified Perceptual Parsing for Scene Understanding},
  booktitle = {Proceedings of the European Conference on Computer Vision (ECCV)},
  pages     = {418--434},
  year      = {2018}
}

@article{zhou2019semantic,
  title={Semantic understanding of scenes through the ADE20K dataset},
  author={Zhou, Bolei and Zhao, Hang and Puig, Xavier and Xiao, Tete and Fidler, Sanja and Barriuso, Adela and Torralba, Antonio},
  journal={International Journal of Computer Vision},
  volume={127},
  pages={302--321},
  year={2019}
}

@inproceedings{Lin2014MicrosoftCOCO,
  author    = {Tsung-Yi Lin and Michael Maire and Serge Belongie and James Hays and Pietro Perona and Deva Ramanan and Piotr Dollár and C. Lawrence Zitnick},
  title     = {Microsoft COCO: Common Objects in Context},
  booktitle = {Computer Vision--ECCV 2014: 13th European Conference, Zurich, Switzerland, September 6-12, 2014, Proceedings, Part V},
  pages     = {740--755},
  year      = {2014},
  publisher = {Springer International Publishing},
}

@inproceedings{he2017mask,
  author    = {Kaiming He and Georgia Gkioxari and Piotr Dollár and Ross Girshick},
  title     = {Mask R-CNN},
  booktitle = {Proceedings of the IEEE International Conference on Computer Vision (ICCV)},
  pages     = {2961--2969},
  year      = {2017}
}

@inproceedings{touvron2021training,
  title={Training data-efficient image transformers \& distillation through attention},
  author={Touvron, Hugo and Cord, Matthieu and Douze, Matthijs and Massa, Francisco and Sablayrolles, Alexandre and Jégou, Hervé},
  booktitle={International Conference on Machine Learning},
  pages={10347--10357},
  year={2021},
  organization={PMLR}
}

@inproceedings{liu2021swin,
  title={Swin transformer: Hierarchical vision transformer using shifted windows},
  author={Liu, Ze and Lin, Yutong and Cao, Yue and Hu, Han and Wei, Yixuan and Zhang, Zheng and Lin, Stephen and Guo, Baining},
  booktitle={Proceedings of the IEEE/CVF International Conference on Computer Vision},
  pages={10012--10022},
  year={2021}
}

@inproceedings{dai2021coatnet,
  title={Coatnet: Marrying convolution and attention for all data sizes},
  author={Dai, Zihang and Liu, Hanxiao and Le, Quoc V. and Tan, Mingxing},
  booktitle={Advances in Neural Information Processing Systems},
  volume={34},
  pages={3965--3977},
  year={2021}
}

@inproceedings{chen2021crossvit,
  title={Crossvit: Cross-attention multi-scale vision transformer for image classification},
  author={Chen, Chun-Fu Richard and Fan, Quanfu and Panda, Rameswar},
  booktitle={Proceedings of the IEEE/CVF International Conference on Computer Vision},
  pages={357--366},
  year={2021}
}

@article{pei2024efficientvmamba,
  title={EfficientVMamba: Atrous selective scan for light weight visual mamba},
  author={Pei, Xiaohuan and Huang, Tao and Xu, Chang},
  journal={arXiv preprint arXiv:2403.09977},
  year={2024}
}

@inproceedings{nguyen2022s4nd,
  title={S4ND: Modeling images and videos as multidimensional signals with state spaces},
  author={Nguyen, Eric and Goel, Karan and Gu, Albert and Downs, Gordon and Shah, Preey and Dao, Tri and Baccus, Stephen and Ré, Christopher},
  booktitle={Advances in Neural Information Processing Systems},
  volume={35},
  pages={2846--2861},
  year={2022}
}

@article{draelos2020use,
  title={Use HiResCAM instead of Grad-CAM for faithful explanations of convolutional neural networks},
  author={Draelos, Rachel Lea and Carin, Lawrence},
  journal={arXiv preprint arXiv:2011.08891},
  year={2020}
}

@article{xia2017aid,
  title={AID: A benchmark data set for performance evaluation of aerial scene classification},
  author={Xia, Gui-Song and Hu, Jingwen and Hu, Fan and Shi, Baoguang and Bai, Xiang and Zhong, Yanfei and Zhang, Liangpei and Lu, Xiaoqiang},
  journal={IEEE Transactions on Geoscience and Remote Sensing},
  volume={55},
  number={7},
  pages={3965--3981},
  year={2017},
  publisher={IEEE}
}

@article{darcet2023vision,
  title={Vision transformers need registers},
  author={Darcet, Timoth{\'e}e and Oquab, Maxime and Mairal, Julien and Bojanowski, Piotr},
  journal={arXiv preprint arXiv:2309.16588},
  year={2023}
}

@inproceedings{gu2022s4,
  title     = {Efficiently Modeling Long Sequences with Structured State Spaces},
  author    = {Gu, Albert and Goel, Karan and R{\'e}, Christopher},
  booktitle = {International Conference on Learning Representations},
  year      = {2022},
  url       = {https://arxiv.org/abs/2111.00396}
}

@inproceedings{wu2021rethinking,
  title     = {Rethinking and Improving Relative Position Encoding 
               for Vision Transformer},
  author    = {Wu, Kan and Peng, Houwen and Chen, Minghao and 
               Fu, Jianlong and Chao, Hongyang},
  booktitle = {Proceedings of the IEEE/CVF International Conference 
               on Computer Vision},
  pages     = {10033--10041},
  year      = {2021}
}

@inproceedings{hu2018squeeze,
  title     = {Squeeze-and-Excitation Networks},
  author    = {Hu, Jie and Shen, Li and Sun, Gang},
  booktitle = {Proceedings of the IEEE/CVF Conference on Computer 
               Vision and Pattern Recognition},
  pages     = {7132--7141},
  year      = {2018}
}

@inproceedings{woo2018cbam,
  title     = {{CBAM}: Convolutional Block Attention Module},
  author    = {Woo, Sanghyun and Park, Jongchan and Lee, Joon-Young 
               and Kweon, In So},
  booktitle = {Proceedings of the European Conference on Computer 
               Vision},
  pages     = {3--19},
  year      = {2018}
}

@article{haruna2025exploring,
  title={Exploring the synergies of hybrid convolutional neural network 
         and Vision Transformer architectures for computer vision: A survey},
  author={Haruna, Yunusa and Qin, Shiyin and Chukkol, Abdulrahman Hamman Adama 
          and Yusuf, Abdulganiyu Abdu and Bello, Isah and Lawan, Adamu},
  journal={Engineering Applications of Artificial Intelligence},
  volume={144},
  pages={110057},
  year={2025},
  publisher={Elsevier},
  doi={10.1016/j.engappai.2025.110057}
}
\end{document}